%% file: 0_main.tex
\documentclass[letterpaper, 10 pt, journal, twoside]{ieeetran}
\IEEEoverridecommandlockouts

\usepackage{graphicx}
\usepackage{amsmath}
\usepackage{upgreek}
\usepackage{bbm}
\usepackage{dsfont}
\usepackage{caption}
\usepackage{subcaption}
\usepackage{color}

\PassOptionsToPackage{dvipsnames}{xcolor}
\PassOptionsToPackage{table}{xcolor}

\usepackage{amssymb}
\usepackage{booktabs}
\usepackage{makecell}
\usepackage[ruled,noend,linesnumbered]{algorithm2e}
\usepackage{titlesec}
\usepackage{wrapfig}
\usepackage{array}
\usepackage{multirow}
\usepackage{colortbl}
\usepackage{tabularx}
\usepackage{macros/mysymbols}

\usepackage{grffile}
\usepackage{xspace}
\usepackage{needspace}
\usepackage{crop}

\usepackage{url}
\usepackage{bm}
\usepackage{breqn}  %
\usepackage{paralist}

\usepackage{pifont}
\newcommand{\task}{mobile pick-and-place\xspace}
\newcommand{\Task}{Mobile pick-and-place\xspace}
\newcommand{\ascURL}{http://adaptiveskillcoordination.github.io}
\newcommand{\ascURLtext}{adaptiveskillcoordination.github.io}

\newcommand{\til}{\raisebox{0.5ex}{\texttildelow}}

\usepackage{array}
\newcolumntype{M}[1]{>{\centering\arraybackslash}m{#1}}
\newcolumntype{N}{@{}m{0pt}@{}}

\usepackage[belowskip=0pt,aboveskip=5pt,font=small]{caption}
\usepackage[belowskip=0pt,aboveskip=0pt,font=small]{subcaption}
\usepackage{xcolor}
\PassOptionsToPackage{hyphens}{url}
\usepackage[hidelinks]{hyperref}
\usepackage{xfrac}
\input{macros/macros.tex}

\usepackage{cite}

\usepackage{hyperref}
\hypersetup{
    colorlinks=true,
    linkcolor=blue,
    filecolor=magenta,      
    citecolor=black,
    pdftitle={2304.00410.pdf},
    pdfpagemode=FullScreen,
}

\usepackage[final]{changes}  %
\definechangesauthor[name={RED}, color=red]{R1}
\definechangesauthor[name={BLUE}, color=blue]{B1}
\setauthormarkup{}

\begin{document}

\title{ASC: Adaptive Skill Coordination for \\ Robotic Mobile Manipulation}
\author
{Naoki Yokoyama$^{1}$, Alex Clegg$^{2}$, Joanne Truong$^{1}$, Eric Undersander$^{2}$, Tsung-Yen Yang$^{2}$, Sergio Arnaud$^{2}$, \\Sehoon Ha$^{1}$, Dhruv Batra$^{1,2}$, Akshara Rai$^{2}$\\
\normalsize{$^{1}$Georgia Institute of Technology,}
\normalsize{$^{2}$FAIR, Meta AI}\\
}

\bstctlcite{IEEEexample:BSTcontrol}

\twocolumn[{
  \renewcommand\twocolumn[1][]{#1}
  \maketitle
  \vspace{-0.5cm}
  \includegraphics[width=\textwidth]{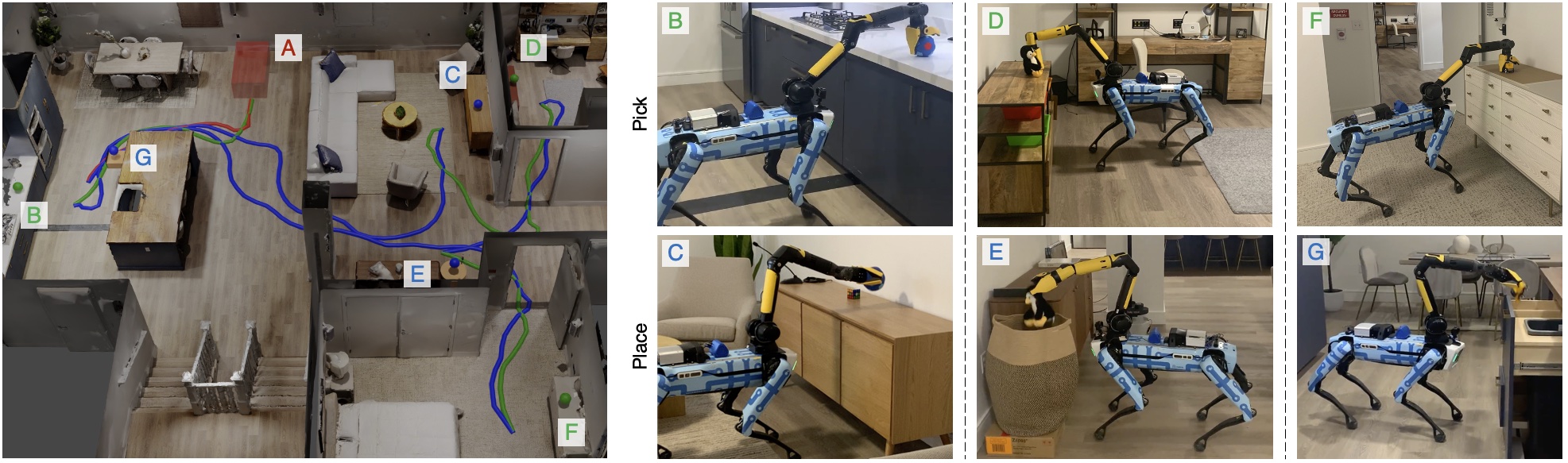}
  \vspace{-0.5cm}
  \captionof{figure} {
    Adaptive Skill Coordination (ASC) is deployed on Spot in a novel environment and tasked with \task, using learned sensor-to-action skills. %
    The robot starts at its dock (red, A), navigates to a pick receptacle (green, B, D, F), searches for and picks an object, navigates to a place receptacle (blue, C, E, G), and places the object at its desired place location, and repeats.
  }
  \label{fig:teaser}
}]

\makeatletter
\let\@oldmakefnmark\@makefnmark
\renewcommand{\@makefnmark}{}

\footnotetext{
Manuscript received: June, 25, 2023; Revised October, 4, 2023; Accepted November, 6, 2023.
}%
\footnotetext{
This paper was recommended for publication by Editor Jens Kober upon evaluation of the Associate Editor and Reviewers' comments.

The Georgia Tech effort was supported in part by ONR YIPs, ARO PECASE, and Korea Evaluation Institute of Industrial Technology (KEIT) funded by the Korea Government (MOTIE) under Grant No.20018216, Development of mobile intelligence SW for autonomous navigation of legged robots in dynamic and atypical environments for real application. JT was supported by an Apple Scholars in AI/ML PhD Fellowship. The views and conclusions are those of the authors and should not be interpreted as representing the U.S. Government, or any sponsor.
} %
\footnotetext{
$^{1}$
NY, JT, SH, and DB are with Georgia Institute of Technology. E-mail:
{\tt\footnotesize{nyokoyama@gatech.edu}}
}
\footnotetext{
$^{2}$
AC, EU, TY, SA, DB, and AR are with Meta AI.
}
\footnotetext{Digital Object Identifier (DOI): see top of this page.}

\let\@makefnmark\@oldmakefnmark
\makeatother

\begin{abstract}
    We present Adaptive Skill Coordination (ASC) -- an approach for accomplishing long-horizon tasks like mobile pick-and-place (\ie, navigating to an object, picking it, navigating to another location, and placing it).
    ASC consists of three components -- (1) a library of basic visuomotor \emph{skills} (navigation, pick, place), (2) a skill \emph{coordination} policy that chooses which skill to use when, and (3) a \emph{corrective} policy that adapts pre-trained skills in out-of-distribution states.
    All components of ASC rely only on onboard visual and proprioceptive sensing, without requiring  detailed maps with obstacle layouts or precise object locations, easing real-world deployment.
    We train ASC in simulated indoor environments, and deploy it \emph{zero-shot} (without any real-world experience or fine-tuning) on the Boston Dynamics Spot robot in eight novel real-world environments (one apartment, one lab, two microkitchens, two lounges, one office space, one outdoor courtyard).
    In rigorous quantitative comparisons in two environments, ASC achieves near-perfect performance (59/60 episodes, or 98\%), while sequentially executing skills succeeds in only 44/60 (73\%) episodes.
    Extensive perturbation experiments show that ASC is robust to hand-off errors, changes in the environment layout, dynamic obstacles (\eg, people), and unexpected disturbances. %
    Supplementary videos at \href{\ascURL}{\ascURLtext}.
\end{abstract}


\IEEEpeerreviewmaketitle

\input{1_intro}

\input{2_related_works}

\input{3_mobile_pick_and_place}

\input{4_adaptive_skill_coordination}
\input{5_training_details}
\input{6_experimental_evaluation}
\input{7_conclusion}

{
    \bibliographystyle{style/IEEEtran}
    \bibliography{bib/strings,bib/main}
}

\clearpage
\input{8_appendix}

\end{document}

%% file: macros/macros.tex
\newcommand{\xhdr}[1]{\vspace{2pt}\textbf{#1}}

\newcolumntype{Y}{>{\centering\arraybackslash}X}
\newcolumntype{P}[1]{\begin{center}>{\arraybackslash}p{#1}\end{center}}

%% file: 1_intro.tex
\section{Introduction}
\IEEEPARstart{W}{e} study `in-the-wild' \task, a subset of the general object rearrangement task~\cite{batra2020rearrangement}.
In this task, a robot is initialized in a home and tasked with moving objects from initial to desired locations, emulating  `tidying-up' (Fig.~\ref{fig:teaser}).
The robot operates entirely using onboard sensors: head- and arm-mounted cameras, proprioceptive joint sensors, and egomotion sensors. 
It navigates to a receptacle with clutter, like the kitchen counter (whose approximate location w.r.t. the robot's starting pose is known), searches for and picks an object (whose name is known but location, pose, or 3D model are \emph{not} known), navigates to the object's desired place receptacle (with known approximate location), places it, and repeats.

The long-horizon nature of this task makes learning challenging:
(1) errors made earlier can limit adequate exploration of states relevant to later parts of the task
(\eg, if the robot takes a wrong turn or gets stuck in navigation, it cannot gather experience for picking the target object);
(2) conflicting goals for different parts of the task can require careful reward tuning that depends on the current stage of the episode
(\eg,~whether the robot is now navigating, picking, or placing);
and (3) learning to complete the initial stages of the task may make learning different skills needed for later stages challenging
(\eg, if the robot learned to avoid obstacles to navigate, it may have difficulty learning to go close to a table for picking).

To tackle these problems, we present Adaptive Skill Coordination (ASC), which consists of three components:
(1) a library of basic visuomotor skills (navigation, pick, and place),
(2) a skill coordination policy that chooses which skills are appropriate to use when,
and (3) a corrective policy that adapts the pre-trained skills in out-of-distribution states. 
All components of ASC use raw high-dimensional observations (\eg, images) and are trained using reinforcement learning (RL) entirely in the Habitat simulator (Habitat-Sim)~\cite{szot2021habitat, habitat19iccv}.
Once trained, it transfers to the real-world `zero-shot', \ie, without any real-world experience.

We deploy ASC on the Boston Dynamics (BD) Spot robot \cite{Spot} and conduct experiments in eight diverse and novel real-world environments (one apartment, one lab, two microkitchens, two lounges, one office space, one outdoor courtyard). %
First, we conduct quantitative comparisons with baselines in two environments -- a fully-furnished $185m^2$ apartment, and a $65m^2$ university lab -- and find that ASC succeeds at a near-perfect rate (59/60 episodes or 98\%), overcoming hardware instabilities, picking failures, and adversarial disturbances like moving obstacles or blocked paths.
In comparison, sequentially executing skills only succeeds in 44/60 (73\%) episodes, due to hand-off errors and inability to recover from disturbances. 
Second, we use ReplicaCAD \cite{szot2021habitat} environments in Habitat-Sim to systematically benchmark several ASC ablations and other baselines over 1500 episodes; we find that ASC's coordination policy plays a significant role in avoiding hand-off errors, and the corrective policy further improves performance over other baselines by adapting to out-of-distribution states. 
Third, we present extensive qualitative evaluations in all eight real-world environments, and demonstrate robustness to dynamic obstacles, adversarial perturbations to the target object, and hardware failures.
Finally, we present a direct quantitative comparison of ASC skills against what the BD API provides `out-of-the-box'.
Overall, ASC assumes limited knowledge of the environment and demonstrates robust `in-the-wild' real-world deployment.

%% file: 2_related_works.tex
\section{Related Works}

\xhdr{Classical task and motion planning (TAMP):}
 Many works study mobile manipulation on humanoid robots~\cite{feng2014optimization, kuindersma2016optimization}, wheeled robots~\cite{heins2021mobile}, and quadrupedal robots~\cite{Combining}. 
 Classical TAMP~\cite{garrett2021integrated} addresses mobile manipulation by forming a task plan given a \textit{planning domain}, and executing a sequence of model-based skills. 
However, it assumes access to privileged information about the test environments such as pre-built detailed maps with obstacles~\cite{heins2021mobile}, precise object locations~\cite{garrett2020online}, or teleoperation~\cite{feng2014optimization, kuindersma2016optimization, Combining}.
ASC uses learned skills and does not depend on a planning domain, detailed maps with obstacle layouts, precise object locations, or teleoperation. 

\xhdr{Modular learning:}
\cite{szot2021habitat, gu2022multi} use a STRIPS task planner~\cite{fikes1971strips} to sequence learned skills; both report hand-off errors as a significant cause of failure, which is a problem we study. %
\cite{li2019hrl4in} and \cite{li2021relmogen} train a coordination policy, but \cite{li2019hrl4in} does not scale to cluttered environments (0\% success as reported by~\cite{li2021relmogen}); \cite{li2021relmogen} assumes access to motion planners and is only demonstrated in simulation.
In the real world, \cite{sun2022fully} learns to pick paper balls in a single room up to $10m^2$ in size. In comparison, we learn to tidy a $185m^2$ fully-furnished unseen real-world apartment. SayCan~\cite{saycan} and TidyBot~\cite{wu2023tidybot} use large language models as task planners and execute basic skills sequentially. However, our experiments show that such sequential skill execution is prone to failure due to hand-off errors. \cite{wu2023tidybot} operates in a small area monitored by third-person cameras used to detect and localize objects. In contrast, we rely solely on onboard sensing, which enables easy deployment to new environments. 

\xhdr{End-to-end learning:} 
\cite{kindle2020wbc, ni2023towards, ehsani2021manipulathor, honerkamp2021kinfeas, wang2020learning, behaviorpriors} learn mobile manipulation end-to-end, but with strong assumptions to simplify the task.
\cite{kindle2020wbc} considers a single corridor with a few obstacles, while \cite{honerkamp2021kinfeas, wang2020learning} do not consider obstacles in the environment.
\cite{ni2023towards, ehsani2021manipulathor} operates in a single simulated room, and assume access to precise object locations. 
\cite{behaviorpriors} operates in a small area monitored by external cameras to localize objects. ASC operates in large, fully-furnished spaces, with only onboard sensing.

%% file: 3_mobile_pick_and_place.tex
\section{Task: \Task}
\begin{figure}[t]
    \vskip5pt
    \centerline{
    \includegraphics[trim=0cm 0cm 0cm 5cm,clip,width=1\columnwidth]{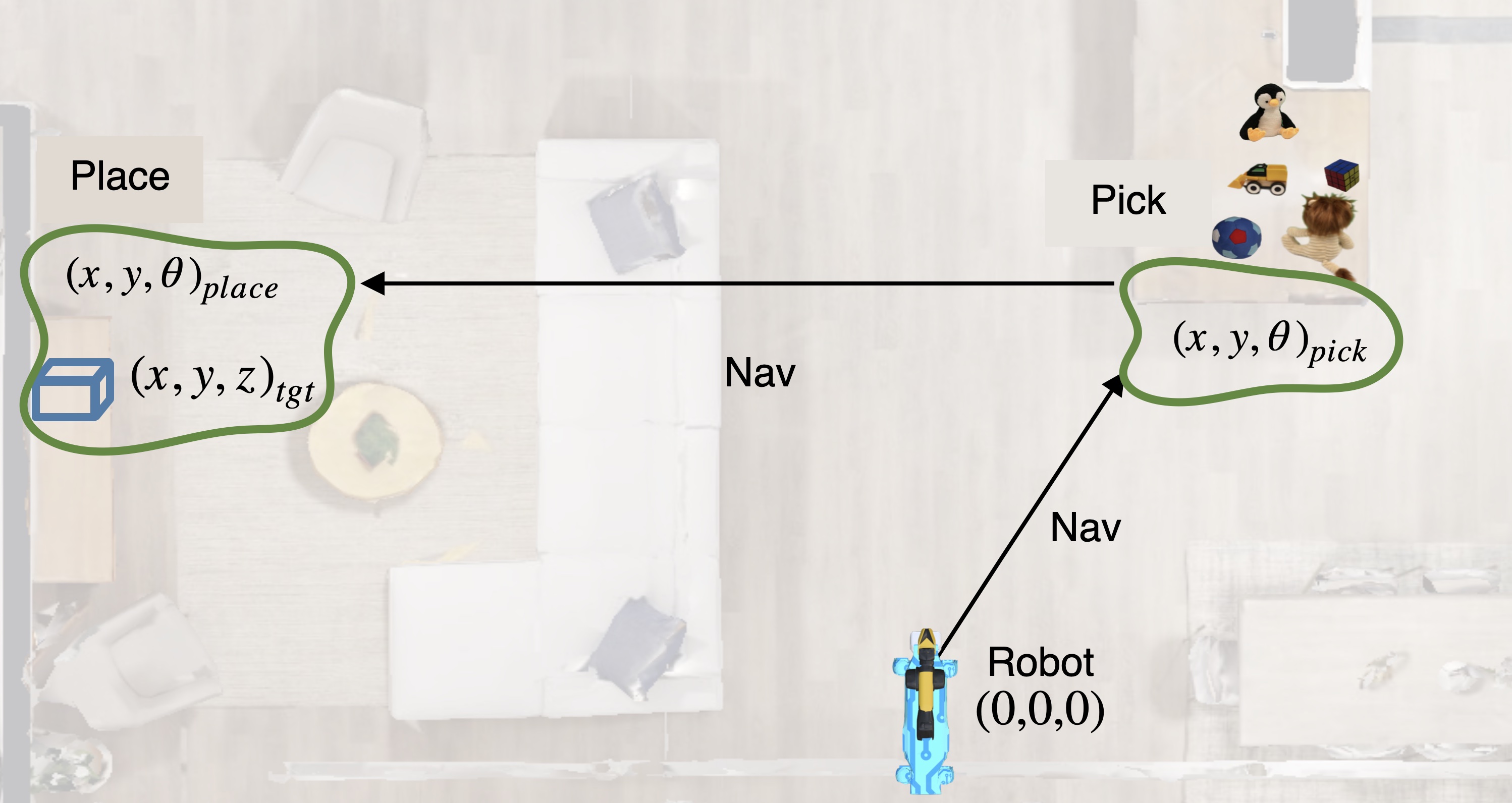}}
    \caption{
        The robot navigates to a receptacle at $(x, y, \theta)_{pick}$, searches for and picks a target object, navigates to the place receptacle, located at $(x, y, \theta)_{place}$, and places the object at the target place location $(x, y, z)_{tgt}$. Precise object locations and a detailed map of the environment with obstacles are not given.
    }
    \label{fig:task_diagram}
    \vspace{0.2cm}
    \centerline{
    \includegraphics[trim=0cm 1cm 0cm 0cm,clip,width=1\columnwidth]
    {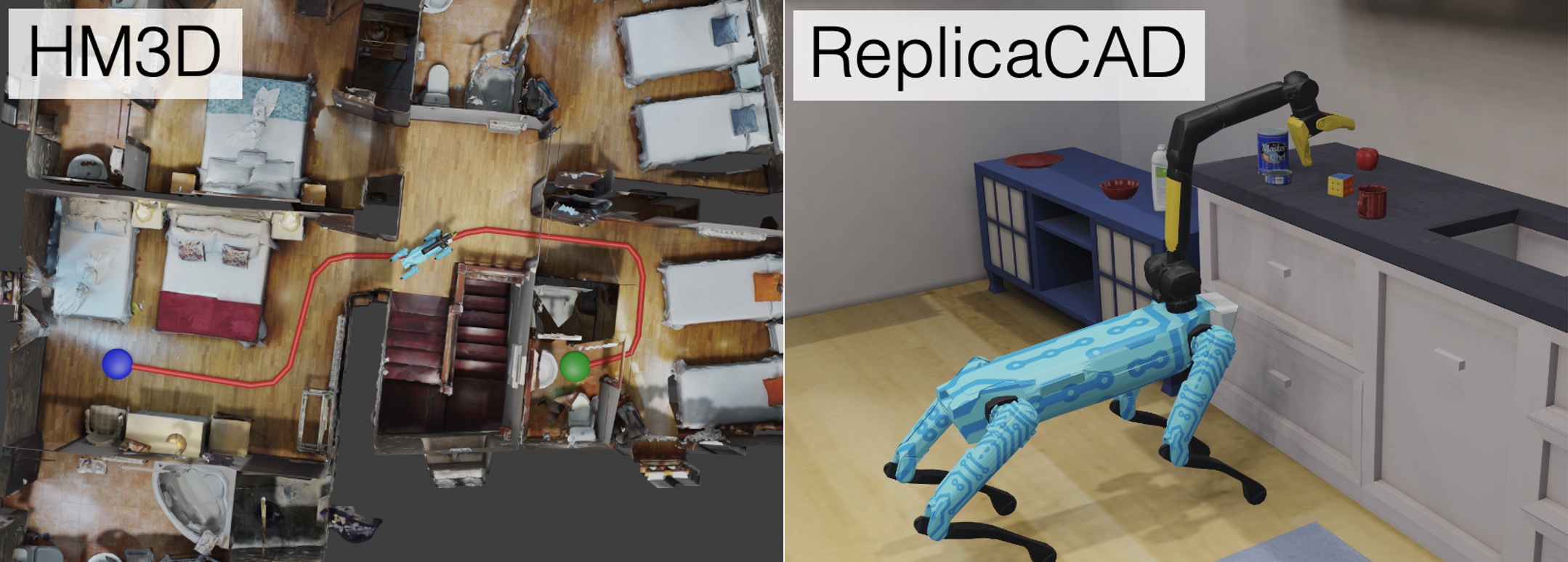}}
    \caption{
        ASC is trained entirely within the Habitat simulator. The HM3D dataset is used to train the navigation skill, while ReplicaCAD is used to train the pick, place, coordination, and corrective policies.
    }
    \label{fig:environment_visualizations}
\end{figure}

The \task task involves finding target objects and placing them in their desired  locations on one of $M$ receptacles, whose approximate locations are known. Receptacles used in our simulation environment include kitchen counters, TV stands, and two types of tables from the ReplicaCAD dataset~\cite{szot2021habitat}. Real-world receptacles include various tables, a sofa, a kitchen counter, a drawer, and a hamper, shown in Fig. \ref{fig:teaser}. An approximate receptacle location is defined as a robot pose from which an object or place location on the receptacle is within 30$^\circ$ of the robot's heading and 0.6m of its position. In simulation, we sample this location using ground-truth object or place locations, and in the real-world, we manually tele-operate the robot to a receptacle and record a robot pose near the approximate center of the receptacle. We follow an episodic coordinate system, established by the robot start pose. In each episode, the robot navigates to the closest receptacle with objects, approximately located at $(x, y, \theta)_{pick}$, where $x$ and $y$ are longitudinal and lateral displacements, and $\theta$ is the relative yaw (Fig.~\ref{fig:task_diagram}).
Upon reaching the receptacle, it searches for and picks a target object, then navigates to the object's desired place receptacle approximately located at $(x, y, \theta)_{place}$, placing the object at $(x, y, z)_{tgt}$.
A detailed map with obstacle layout or precise object locations is not provided.

\xhdr{Robot observations.}
The full observation space (Fig.~\ref{fig:framework_architecture}) consists of:
(1) two depth images from the front of the robot, concatenated to make a single image $I_{front}$;
(2) a depth image from the gripper  $I_{grip}$;
(3) a bounding box image $I_{bbox}$ of the target object, if detected by an object detector;
(4) the joint angles of the robot arm $q_{arm}$;
and (5) an egomotion sensor estimating the robot's relative heading and displacement from its start pose. 
The approximate receptacle location where the object can be picked or placed (\eg, counter, hamper) and the place target are also provided.

%% file: 4_adaptive_skill_coordination.tex
\begin{figure*}[t!]
\centering
    \vskip5pt
    \includegraphics[width=1.0\textwidth, trim=0cm 0cm 0cm 1.5cm, clip]{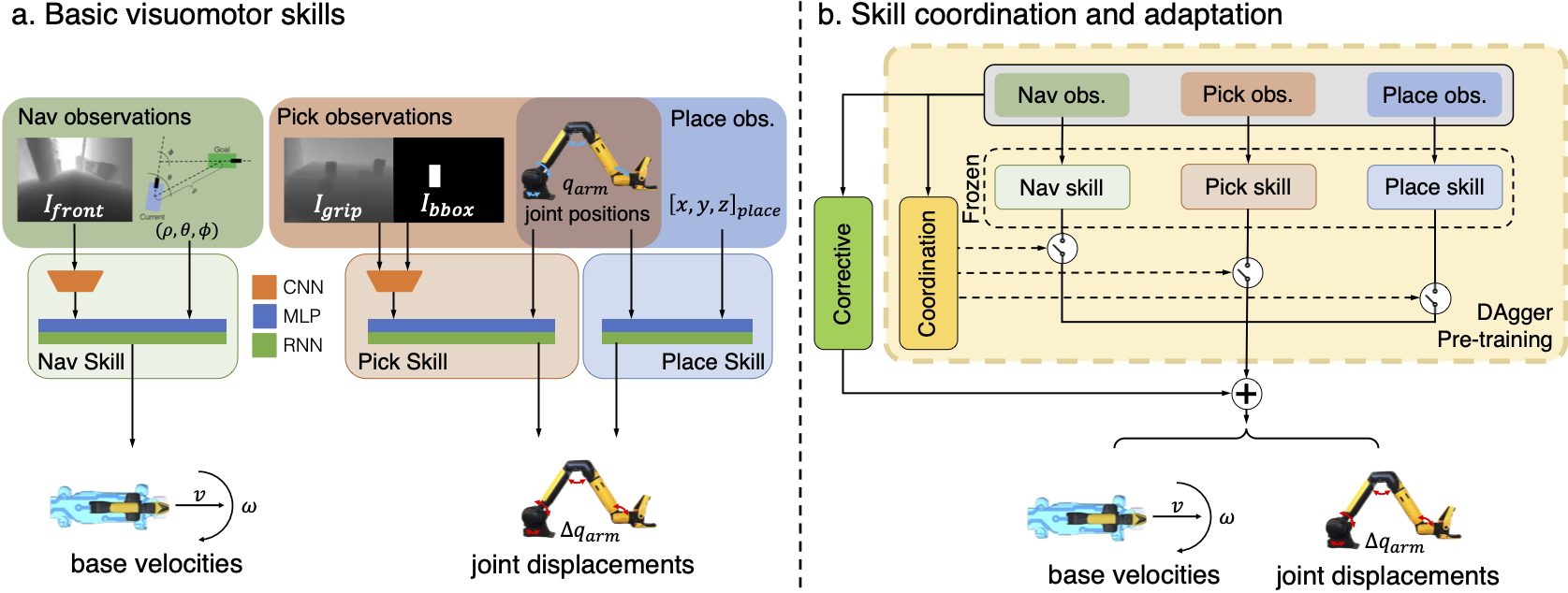}
    \caption{
        Training ASC consists of two steps: (left) First, we train a library of 3 basic visuomotor skills in diverse simulated environments. The skills are trained using RL to achieve the relatively shorter-horizon tasks of navigation, picking and placing, and command robot base velocities and joint position deltas. (right) Next, we train a skill coordination policy that chooses which skills are appropriate to use based on observations, and a corrective policy that adapts the pre-trained skills in out-of-distribution states, for the task of \task. 
    }
    \label{fig:framework_architecture}
\end{figure*}

\newcommand{\dxy}[2]{d({#1},{#2})}
\newcommand{\dtheta}[2]{\Theta({#1},{#2})}
\newcommand{\dist}{\Delta_t}
\newcommand{\distlast}{\Delta_{t-1}}
\newcommand{\p}{\boldsymbol{p}}
\newcommand{\brho}{\boldsymbol{\rho}}

\section{Adaptive Skill Coordination}
Training ASC consists of two steps.
First, we train a library of three basic visuomotor skills: navigation, picking, and placing (Fig.~\ref{fig:framework_architecture}, left).
Next, we train a skill coordination policy that chooses which skills to use when, and a corrective policy that is used instead of the pre-trained skills when out-of-distribution states are perceived (Fig.~\ref{fig:framework_architecture}, right). All policies are trained using RL in simulation.
In the following section, $\dxy{a}{b}$ denotes the geodesic distance (length of the shortest obstacle-free path) between two points $a$ and $b$;
$\dtheta{a}{b}$ is the angular distance between two orientations $a$ and $b$; and
$\mathbb{I}^{condition}=1$ if \textit{condition} is true, and $0$ if not. $\p$ denotes the position and $\brho$ the orientation components of a pose.

\subsection{Basic visuomotor skills}

\xhdr{Navigation skill.}
Spot is initialized in a random location in a scene from the HM3D dataset \cite{ramakrishnan2021hm3d} (Fig.~\ref{fig:environment_visualizations}), and tasked with reaching within 0.3m and 5$^{\circ}$ of a goal $(x, y, \theta)_{goal}$, specified relative to its start pose (similar to PointNav \cite{anderson2018evaluation}). The skill $\pi_{\text{nav}}$ uses the front depth images $I_{front} \in R^{240 \times 212}$, and an egomotion sensor that measures the current pose of the robot's base $(x, y, \theta)^t_{base}$ relative to the start pose, and outputs local forward and angular velocity commands: $v, \omega \sim \pi_{\text{nav}}(I_{front}, (x, y, \theta)^t_{base}, (x, y, \theta)_{goal}) $,
where $v$ and $\omega$ have ranges ${[-0.5, 0.5]} \sfrac{m}{s}$ and $[-30, 30]\sfrac{^\circ}{s}$.
 $\p_{base}^t$ denotes the 2D position of the robot base $(x, y)^t_{base}$ at time $t$, and $\p_{goal}$ denotes $(x,y)_{goal}$. 
$\dist^{\text{$nav$}}$ denotes a base-distance-to-goal that incorporates both  location and heading: 
\newcommand{\yawdiff}{\dtheta{\theta_{base}^t}{\theta_{goal}}}
\begin{equation*}
\begin{split}
    \dist^{\text{$nav$}} = \dxy{\p_{\text{$base$}}^t}{\p_{goal}} 
    \quad\quad\quad\quad\quad\quad\quad\quad 
    \quad\quad\quad\quad\quad\quad \\
    + 0.1 \cdot
    \begin{cases}
        \yawdiff, \quad \text{if }
        \dxy{\p_{base}^t}{\p_{goal}} <1\text{m} \\
        c,
        \quad \quad \quad \quad \quad  
        \quad \ \  
        \text{otherwise}
    \end{cases}
\end{split}
\end{equation*}
where $c=3.14$ is a constant; $\dist^{\text{$nav$}}$ only encourages the robot to orient itself to $\theta_{goal}$ when it is within 1m of the goal location.
The navigation reward also gives a constant slack penalty of $0.01$ (for faster task completion), penalizes collision, backward movement, and gives a terminal reward if successful:
\vspace{-0.1cm}
\begin{dmath*}
    r_{\text{nav},t} = (\distlast^{\text{$nav$}} - \dist^{\text{$nav$}}) - 0.003\ \mathbb{I}^{collision} - 0.003\ \mathbb{I}^{backward} + 5\ \mathbb{I}^{success} - 0.01
\end{dmath*}
\vspace{-0.1cm}
\newcommand{\xyzobj}{\p_{obj}}
\newcommand{\xyzgrip}{\p_{grip}^t}
\newcommand{\eeToObjDist}{\dxy{\xyzgrip}{\xyzobj}}
\newcommand{\objectAngle}{\dtheta{\brho_{grip}^t}{\brho_{obj}^t}}
\newcommand{\distToGoal}{|\eeToObjDist - 0.4| + \objectAngle}

\xhdr{Pick skill.}
The goal of $\pi_{pick}$ is to: a) search for the target object, b) center it in the gripper camera view, and then c) pick it.
For c), we use the BD grasp API which takes as input a pixel located on the target object.
However, our experiments (see Sec.~\ref{sec:bd_comparison} and videos at \href{\ascURL}{\ascURLtext}) show that the BD grasp API often fails if the target object is far from the gripper ($>$\til0.8m), or occluded by other objects; thus, a) and b) are critical, and are the focus of $\pi_{pick}$.
In simulation, Spot is trained to reach its arm towards an object on one of the receptacles in the ReplicaCAD scene (Fig.~\ref{fig:environment_visualizations}), starting from random arm joint positions.
Multiple distractor objects are placed on the receptacle to simulate clutter and occlusion. 
$\pi_{pick}$ receives as input the current joint positions of the arm $q_{arm}$, and two images: the depth image from the gripper camera $I_{grip} \in R^{240 \times 228}$, and a binary mask indicating which pixels are within the bounding box of the target object $I_{bbox} \in R^{240 \times 228}$.
If the target object is not visible, $I_{bbox}$ is all zeros (no bounding box detected). 
The formulation of $I_{bbox}$ allows the pick skill to generalize to novel objects at test-time, as long as an object detector can provide its bounding box.
$\pi_{pick}$ outputs desired joint angle displacements $\Delta q$ relative to the current arm joint positions $q_{arm}:
    \Delta q \sim \pi_{pick} (I_{grip}, I_{bbox}, q_{arm})$.
Let $\xyzgrip$, $\xyzobj$ denote the 3D positions of the gripper and the object; 
$\brho_{grip}^t$ denote the gripper camera's principle axis;
and $\brho_{obj}^t$ denote the unit vector from the gripper camera's center to the target object's center (all are known in sim during training). 
Analogous to navigation, we define a gripper-distance-to-goal, where the goal is to have the gripper $0.4$m away from the target object and pointing directly at it:
$\dist^{\text{$pick$}} = \distToGoal$.
The reward for $\pi_{pick}$ encourages reducing gripper-distance-to-goal over time, penalizes picking the wrong object, and uses a slack penalty of $0.01$ and a terminal success reward:
\begin{equation*}
    r_{\text{$pick,$}t} = 20\ (\dist^{\text{$pick$}} - \distlast^{\text{$pick$}}) - 5\ \mathbb{I}^{wrong\_pick} + 10\ \mathbb{I}^{success} - 0.01
\end{equation*}
The BD grasp API is called if $\objectAngle<15^\circ$ (\ie, object is centered) and the object is $0.3-0.75$m away from the gripper for 4 consecutive time steps ($2$ seconds).
To determine if these criteria are met, we use the gripper depth image and the center of the detected bounding box to calculate the distance and $\objectAngle$.
In simulation, the object is  kinematically attached to the gripper, avoiding expensive contact-rich simulation, following  \cite{truong2022rethinking}. 
Note, $\pi_{pick}$ is not given the precise location of the target object (easing real-world deployment), and searches for it with $I_{grip}$ and $I_{bbox}$. %

\newcommand{\xyztgt}{\p_{tgt}}
\newcommand{\eeToTgtDist}{\dxy{\xyzgrip}{\xyztgt}}
\newcommand{\downangle}{\dtheta{\brho_{grip}^t}{-\vec{z}}}
\xhdr{Place skill.}
To train the place skill $\pi_{place}$, Spot is spawned in front of one of the receptacles in ReplicaCAD (Fig.~\ref{fig:environment_visualizations}), while holding an object in its gripper.
The object must be placed within 0.1m of its desired place location $\p_{tgt} = (x, y, z)_{tgt}$, located on the receptacle. 
One limitation of Spot's embodiment is that the gripper camera is blocked by the gripped object, and front cameras are often blocked by receptacles when Spot is standing close to them; thus, the place skill does not take visual inputs. 
The output of $\pi_{place}$ are the desired joint angle displacements $\Delta q$ relative to the current arm joint positions $q_{arm}$:
$\Delta q \sim \pi_{place} (q_{arm}, (x, y, z)_{tgt})$.
The reward encourages the arm to reach the place target using a distance term, $\dist^{\text{$place$}} = \eeToTgtDist$. It also encourages the gripper to change its orientation to point downwards soon before the object should be released ($<$0.4m from $(x, y, z)_{tgt}$), so that the object can fall naturally upon opening the gripper, using 
\begin{equation*}
    \dist^{\text{$place$}, \theta} = 
    \begin{cases}
        \downangle,
        & \text{if }
        \eeToTgtDist < 0.4\text{m} \\
        3.14,
        & \text{otherwise}
    \end{cases}
\end{equation*}
where $\downangle$ is the angle between the forward axis of the gripper and the negative $\vec{z}$ axis, and $\p_{grip}^{t}$ is the current gripper position. The reward also penalizes collisions with the environment, and uses a success reward and a slack penalty:
\vspace{-0.1cm}
\begin{dmath*}
    r_{\text{$place,$}t} = 5\ (\distlast^{\text{$place$}} - \dist^{\text{$place$}}) - 0.003\ \dist^{\text{$place$}, \theta} - 0.03\ 25\ \mathbb{I}^{success} - 0.01
\end{dmath*}
\vspace{-0.1cm}

In principle, it is possible to replace this place skill with a motion planner that reaches a target end-effector location.
However, learning the place skill enables training skill coordination and correction with the place skill in-the-loop.

\subsection{Skill coordination and correction}\label{subsec:skill_coor_and_corr}
In the second stage, we train a mixture-of-experts coordination policy $\pi^{coord}$ that activates the appropriate (pre-trained and frozen) skills, depending on observations. 
However, since the basic skills have never seen the full \task task, they may have difficulty generalizing to out-of-distribution states observed in the course of this long-horizon task. 
To handle such cases, we train a corrective policy $\pi^{corr}$ that adapts to such out-of-distribution states. 

\newcommand{\coordpolicy}{\pi^{coord}}
\newcommand{\coordaction}{a_{coord}}
\xhdr{Skill coordination.}
Given a library of $K$ pre-trained skills $\Pi = \{\pi_1, \pi_2 \cdots \pi_K \}$, a mixture-of-experts coordination policy $\coordpolicy$ learns to coordinate skills based on observations.
Let $z_i \sim \pi^{z}(\pi_i|o)$ be the outputs of a learned gating network $\pi^{z}$ representing the probability that skill $\pi_i$ is selected. We threshold the \textit{tanh}-activated output of $\pi^{z}$ around 0 (\textit{i.e.,} output $< 0$ is $z_i = 0$, and output $> 0$ is $z_i = 1$) to obtain a binary selection variable per skill, \ie,~$z_i \in \{0, 1 \}$, with $\pi_i$ being selected if $z_i = 1$. 
Multiple or no skills can be selected at the same time, \ie, $0 \leq \sum_i z_i \leq K$. 
This allows the robot to move its base at the same time as reaching for an object, or operate using only actions from the corrective policy, if appropriate.

Let $a_i$ be the action sampled from skill $\pi_i$ given the relevant subset of observations $o_i$. 
Then $\coordaction \sim \coordpolicy(o)$ is: %
\vspace{-0.1cm}
\begin{equation*}
    \coordaction = \sum_{i=1}^K z_i \cdot a_i,
    \ \ \text{where}\  z_i \sim \pi^{z}(\pi_i|o),\  a_i \sim \pi_i(o_i).
    \vspace{-0.1cm}
\end{equation*}
The action spaces for the basic skills can differ; for example, the pick and place skills control the robot's arm, while the navigation skill controls its base. 
 For skills with complementary action spaces $a_{arm}$ and $a_{base}$, we create a combined action space $a_{robot} = a_{arm} \oplus a_{base}$, which is a concatenation of the individual skill actions. 
 For skills that share action spaces (like pick and place), we add an additional constraint that only one of the skills can be activated at a time. 
 Specifically, for a subset $a_{sub} \in \{a_{arm}, a_{base} \}$, for all skills $\pi_j$ whose action space $a_j \in a_{sub}$, we enforce that $\sum_j z_j = 1$
by only selecting the skill that has the highest \mbox{\textit{tanh-activation}}.
This creates a hierarchical MoE~\cite{jordan1994hierarchical}, separating the regions where skills with shared actions act, while sharing the space between complementary skills. 
Intuitively, this separates pick and place, while allowing navigation and pick or navigation and place to operate together.
In the future, it is easy to extend to several skills that share action spaces, for example multiple place skills for different objects, or opening/closing skills, and a learned hierarchical MoE that chooses between them. 
 
\xhdr{Skill correction.} 
Since the skills are trained on simpler tasks (\eg, isolated navigation, picking, placing), there are likely to be states in \task where they perform poorly. 
We adapt to such states using a corrective policy $\pi^{corr}$, without losing the knowledge gained during the first stage of training.
We use the output of the gating network $\pi^{z}$ to indicate if a skill should be adapted or not. 
Intuitively, $\pi^{z}$ picks skills that perform well given observation $o$ and need not be adapted.
Conversely, it does not select skills that perform poorly in the current state, and hence should be adapted.
Thus, we adapt the action sub-space(s) \textit{unused} by $\pi^{z}$ using corrective actions from the corrective policy $\pi^{corr}$. Note that $\pi^{corr}$ adapts to states that are out-of-distribution for the pre-trained skills, but are in-distribution for $\pi^{corr}$ itself, which is trained on the full \task task.

If all skills $\pi_j$ with $a_j \in a_{sub}$ are not selected, the adaptive action $a_{adapt} \in a_{sub}$ using $a_{corr} \sim \pi_{corr}(o)$ is:
\begin{equation*}
    a_{adapt}(o) = 
    \begin{cases}
        0, & \text{if } \sum_j z_j \geq 1, \quad z_j \sim \pi^{z}(\pi_j|o) \\
        a_{corr}, & \text{otherwise}
    \end{cases}
\end{equation*}
The final action, $a_{ASC} = \coordaction + a_{adapt}$, is a sum of the coordination and adaptive actions.
$a_{ASC}$ consists of both base and arm actions $(v, \omega, \Delta q)$.
The input to both the coordination and corrective policies is the superset of observations used to train the skills (Fig.~\ref{fig:framework_architecture}): \mbox{$o = \{I_{front},I_{grip}, I_{bbox}, (x, y, \theta)_{goal}, q_{arm}, (x, y, z)_{tgt}\}$}.
$(x, y, \theta)_{goal}$ is the location of the desired pick or place receptacle, depending on if the robot has picked the object yet. 
To process the image observations ($I_{front}, I_{grip}, I_{bbox}$), we reuse the trained visual encoders from the basic skills.
This enables extraction of useful visual features without re-training the visual encoders.
Giving the corrective and coordination policies a superset of visual inputs also enables more efficient \task;
for example, if the coordination policy sees the target object detected in $I_{bbox}$ before fully reaching $(x, y, \theta)_{pick}$, it can terminate navigation early and initiate picking.

\xhdr{Reward.}
To train the coordination and corrective skills, the robot is tasked with moving one object from one receptacle to a target place location on a different receptacle, using:

\vspace{-0.075cm}
\begin{dmath*}
    r_t = 10\ \mathbb{I}^{\text{success}} - 0.03\ \mathbb{I}^{\text{collision}}  - 0.03\ \mathbb{I}^{\text{backward}} - 0.003\ e_t - 0.01
    \label{eq:sparse_reward}
\end{dmath*}
\vspace{-0.075cm}
where $e_t$ is the sum-of-squares (magnitude) of the actions.

%% file: 5_training_details.tex
\section{Training Details}

All policies in ASC are trained in Habitat-Sim using DDPPO~\cite{wijmans2020ddppo}.
A sparse reward function for training coordination and correction, as described in the previous section, makes RL challenging due to a large state-action space where most action sequences result in suboptimal rewards. To overcome this, we pre-train the coordination policy using the DAgger algorithm~\cite{ross2011reduction} with sequential skill-chaining as a teacher (\textit{Seq-skills} in Sec.~\ref{sec:sim_results}).
$\pi_z$ is trained to activate skills in a pre-defined sequence (navigate, pick, navigate, place) based on the robot's observations. The corrective policy, on the other hand, is not warm-started with pre-training due to the absence of labels. Both the coordination and corrective policies are then fine-tuned using RL, resulting in significantly improved performance compared to sequential skill-chaining (experiments in Sec.~\ref{sec:sim_results}). By only incorporating dense rewards for training basic skills and relying on a simple sparse reward for the full task, we minimize the need for extensive RL reward tuning.

\xhdr{Policy architecture.}
Each policy has a 3-layer convolutional neural network (except the place skill), followed by a 2-layer perceptron and a recurrent GRU~\cite{cho2014gru} layer, and finally a fully-connected layer that parameterizes a diagonal Gaussian action distribution.
Specifically, a policy $\pi$ given observations $o$ outputs actions $a \sim \mathcal{N}(\mu, \sigma \textbf{I})$, where $[\mu, \sigma ] = \pi(o)$. Our real-world experiments use a Razer Blade 14 consumer-grade laptop, with an RTX 3070 Max-Q Mobile GPU and an AMD Ryzen 9 CPU; total inference time is about 5 ms.

%% file: 6_experimental_evaluation.tex
\begin{figure}[t]
    \vskip5pt
    \centerline{\includegraphics[width=1.0\columnwidth]{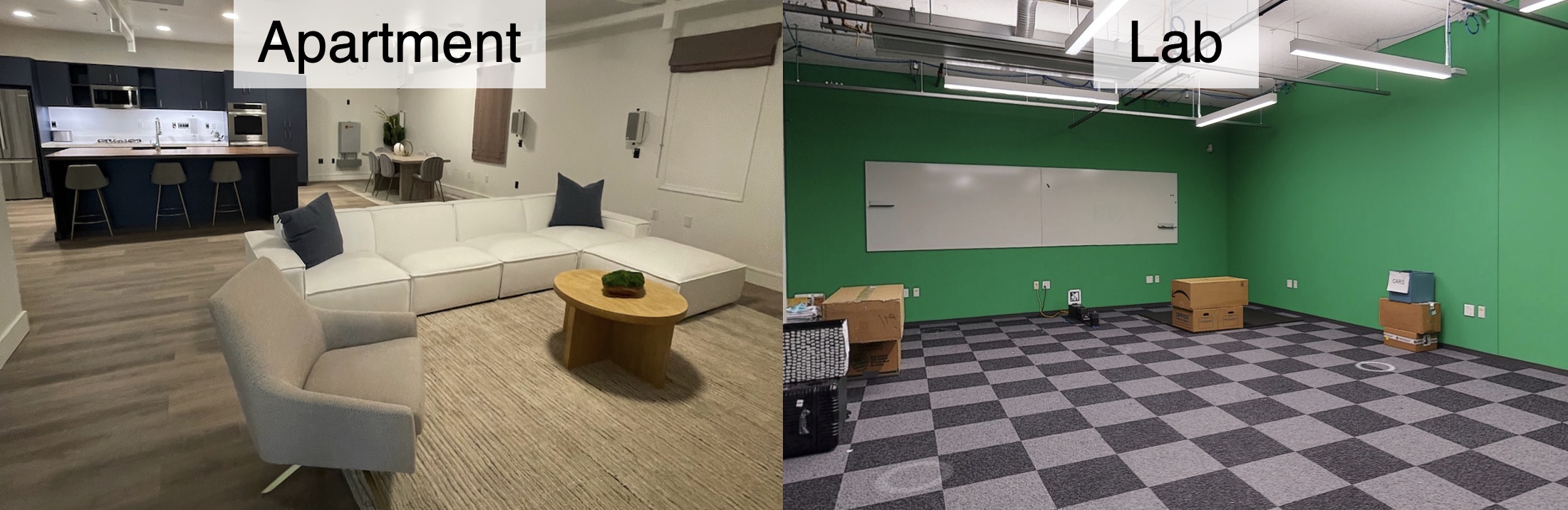}}
    \caption{
        We present quantitative experiments in an \textit{Apartment} and a \textit{Lab}, 2 of the 8 unseen real-world environments ASC is deployed in.
    }
    \label{fig:real_envs}
\end{figure}

\section{Experimental Evaluation}

\xhdr{Real-world environments.}
We deploy ASC on Spot in the real-world, 
and run quantitative experiments in two environments: a $185 m^2$ fully-furnished apartment (\textit{Apartment}) and a $65 m^2$ university lab (\textit{Lab}), shown in Fig.~\ref{fig:real_envs}.
In the \textit{Apartment} the robot needs to take indirect routes to the goal (\eg,~navigate out of a room to get to the goal in the next room), while in the \textit{Lab} the robot needs to navigate a cramped space to avoid collisions.
Detailed information about these environments, like obstacle layout, size and shape of receptacles, or precise object locations, is not known apriori.
All hyperparameters for all approaches are tuned to maximize their performance in a held-out simulation environment.
During real-world deployment, these hyperparameters are kept constant.

\xhdr{Robot control.}
In simulation, we use an articulated Spot URDF model provided by BD~\cite{Spot}.
Following~\cite{truong2022rethinking} for sim-to-real transfer, we use an approximate kinematic simulation of the robot, and depend on low-level hardware controllers in the real world via the BD API.
For navigation,  commanded velocities are integrated using a fixed timestep of 0.5 seconds (2 Hz), and the robot is teleported to the resultant waypoint, if there are no collisions.
Otherwise, the policy receives a collision penalty during training and no movement occurs.
Likewise, the arm is moved to the desired  joint positions if it would not cause collision.
In the real world, the learned policy outputs actions at 2 Hz, which are sent to the BD API, and sensor observations are received from the robot at a frequency of 12 Hz. We also use the BD API to localize the robot and obtain its current pose relative to its start pose. Note that the BD API does not check environment collisions with the arm, and base obstacle avoidance is turned off to allow the robot to move closer to receptacles.
Hence, all collision avoidance behavior is learned entirely  in simulation.

\begin{figure*}
    \vskip5pt
    \centering
    \begin{subfigure}[b]{\textwidth}
        \centering
        \includegraphics[width=\textwidth]{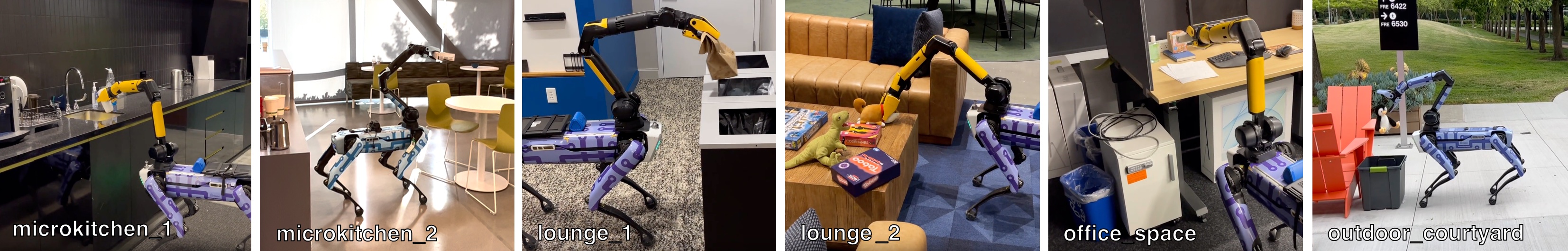}\\
        \vspace{1mm}
        \label{fig:env_layout}
    \end{subfigure}
    \caption{
       ASC is deployed in eight real environments (including \textit{Apartment} and \textit{Lab}), showing in-the-wild \task capabilities.
    }
\label{fig:in-the-wild}
\end{figure*}
\subsection{Quantitative experiments in the real world}
ASC is tasked with rearranging 30 objects (ranging from a plush penguin/lion/ball, three different toy cars, and a Rubik's cube) in the \textit{Apartment} and \textit{Lab}.
Mask R-CNN~\cite{he2017mask} is used for object detection.
The robot navigates to the closest pick receptacle, searches for and picks an object, navigates to that object's desired place receptacle, and places the object at its desired place location (Fig.~\ref{fig:teaser}).
The process is repeated 30 times, to allow the robot to attempt to rearrange each object once.
We compare ASC to:
\newcommand{\nocorrective}{ASC-NoCorrective}
\newcommand{\nodagger}{ASC-NoDAgger}
\newcommand{\finetuning}{ASC-NoCorrective-Finetune}
\newcommand{\seqskillsmap}{Seq-skills-premapping}
\begin{itemize}
    \item \nocorrective: ASC with only coordination and no corrective policy, to study the role of the corrective policy.
    \item Seq-skills: A skill-chaining baseline that sequentially executes each skill (navigation, pick, navigation, place).
    \item \seqskillsmap: Same as Seq-skills, but the learned navigation skill is replaced with a traditional mapping-based approach.
    Unlike all other approaches, \seqskillsmap\ receives a pre-built privileged, detailed map of the environment, which is used by a graph-based navigation API provided by BD.
    This baseline can fail if the layout of the environment changes after mapping~\cite{spotgraphnav}, which learned navigation is robust to.
\end{itemize}

\begin{figure*}
    \vskip5pt
    \centering
    \begin{subfigure}[b]{\textwidth}
        \centering
        \includegraphics[width=\textwidth]{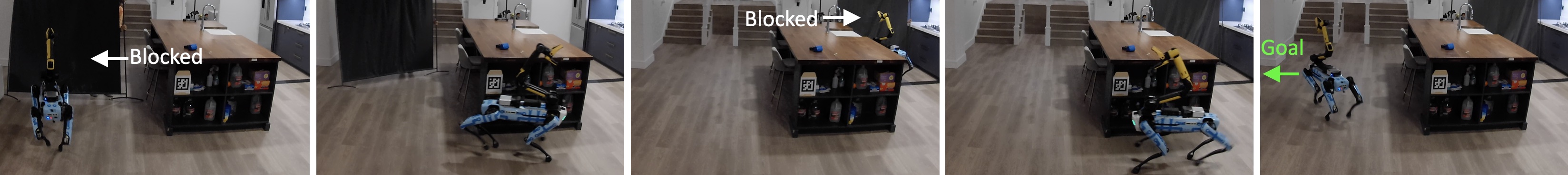}\\
        \vspace{1mm}
        \label{fig:env_layout}
    \end{subfigure}
    \begin{subfigure}[b]{\textwidth}
        \centering
        \includegraphics[width=\textwidth]{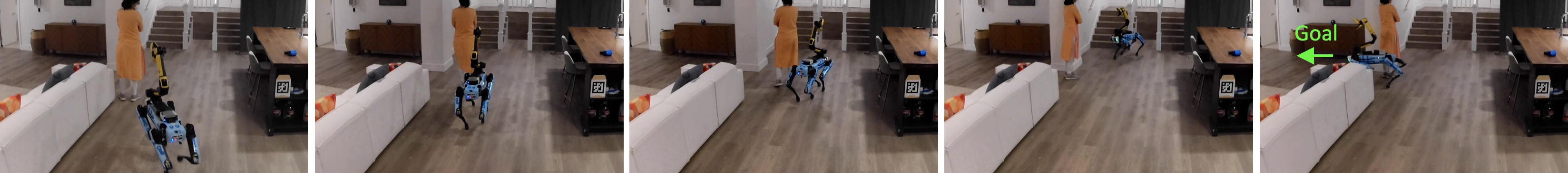}\\
        \vspace{1mm}
    \end{subfigure}
    \begin{subfigure}[b]{\textwidth}
        \centering
        \includegraphics[width=\textwidth]{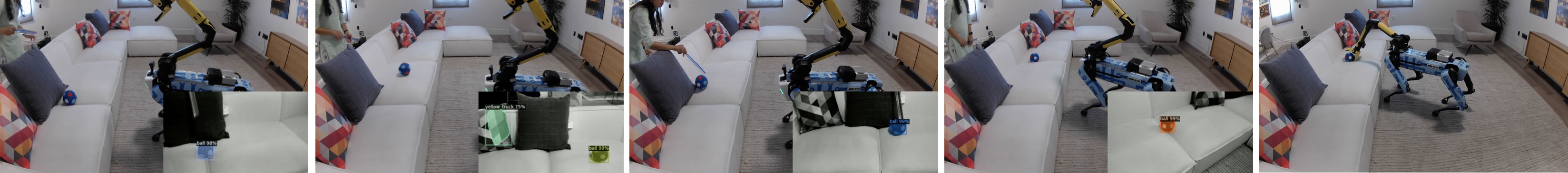}\\
        \label{fig:moving_obj}
    \end{subfigure}
    \caption{
       ASC is robust to several types of real-world disturbances (see video at \href{\ascURL}{\ascURLtext}). \textit{Top}: ASC can reroute the robot upon sudden changes to the environment layout blocking its path, until it successfully finds a path to the goal, even if that path was previously blocked. \textit{Bottom}: Since ASC does not take object positions as input, it is also robust to perturbations like moving objects. %
    }
\label{fig:perturbation}
\end{figure*}

\xhdr{Real-world performance.} We measure performance as successful \task episodes, \ie, the number of objects that the robot successfully picked and placed within $0.1$m of their target place locations over a total of 60 episodes -- 30 in \textit{Apartment} and 30 in \textit{Lab}, over three runs of 10 episodes each.
In \textit{Apartment}, each run of 10 \task episodes took about 10-15 minutes and the robot moved approximately $150$m, while in the \textit{Lab} it took about 8-10 minutes and moved about $135$m.
The robot was completely autonomous throughout all experiments, needing no human intervention. 
Tab.~\ref{table:success_table} shows a quantitative comparison.
ASC successfully rearranged $29/30$ objects in \textit{Apartment}, and $30/30$ objects in \textit{Lab}, missing one object due to the BD grasp API failing to pick a toy car.
In comparison, \nocorrective\ rearranged $27/30$ in \textit{Apartment} and $27/30$ in \textit{Lab}, failing due to the inability of $\pi_{nav}$ to move close enough to the pick or place receptacle. $\pi_{nav}$ is trained to avoid obstacles, like receptacles, and hence moving close is out-of-distribution and requires the corrective policy.
Seq-skills rearranged $23/30$ objects in \textit{Apartment} and $21/30$ objects in \textit{Lab}, predominantly failing due to hand-off errors when $\pi_{nav}$ stopped the robot in a suboptimal position for subsequent picking or placing.
\seqskillsmap\ rearranged $22/30$ objects in the \textit{Apartment}, and $21/30$ in the \textit{Lab}, also failing due to hand-off errors, and once due to a navigation error, where the robot became stuck and could not navigate to the target location.
\seqskillsmap\ performed similar to Seq-skills, indicating that hand-off errors were not eliminated by utilizing a classical map-based navigation method instead of learned navigation.
Both performed worse than ASC, which is robust to hand-off errors and re-invoked the navigation skill or used the corrective policy to move the robot if needed.

\definecolor{Gray}{gray}{0.9}
\begin{table}[t]
    \centering
    \resizebox{1.0\columnwidth}{!}{
        \begin{tabular}{ccccc}
\toprule
Row & & Apartment (real) & Lab (real) & Simulation \\
\# & & 30 episodes & 30 episodes & 1500 episodes \\
\midrule
\rowcolor{Gray}
1 & ASC & \textbf{96.7\%\tiny{$\pm$}3.3} & \textbf{100.0\%\tiny{$\pm$}0.0} & \textbf{94.9\%\tiny{$\pm$}0.6} \\
\rowcolor{Gray}
2 & \nocorrective & 90.0\%\tiny{$\pm$}5.5 & 90.0\%\tiny{$\pm$}5.5 & 92.3\%\tiny{$\pm$}0.7 \\
3 & Seq-skills & 76.7\%\tiny{$\pm$}7.7 & 70.0\%\tiny{$\pm$}8.4 & 87.8\%\tiny{$\pm$}0.8 \\
4 & \seqskillsmap & 73.3\%\tiny{$\pm$}8.1 & 70.0\%\tiny{$\pm$}8.4 & - \\
\rowcolor{Gray}
5 & \finetuning & - & - & 43.9\%\tiny{$\pm$}1.3 \\
\rowcolor{Gray}
6 & \nodagger & - & - & 5.6\%\tiny{$\pm$}0.6 \\
7 & Mono-RL & - & - & 35.8\%\tiny{$\pm$}1.2 \\

\bottomrule
        \end{tabular}
    }

\caption{
    We evaluate ASC on \task in \textit{Apartment}, \textit{Lab}, and simulation, comparing ASC against various baselines and ablations to highlight the significance of each component of ASC.
}
\label{table:success_table}
\vspace{-0.25cm}
\end{table}

\xhdr{In-the-wild \task.}
We also deploy ASC in 6 additional real-world environments (see Fig.~\ref{fig:in-the-wild}), highlighting its robustness at solving \task in-the-wild.
We use Owl-ViT~\cite{owlvit} to detect a wide range of objects.
Videos can be found at \href{\ascURL}{\ascURLtext}.

\subsection{\Task in simulation}
\label{sec:sim_results}
We perform additional comparisons in simulation, including the following additional baselines and ablations of ASC:
\begin{itemize}
    \item Fine-tuning: An ablation of ASC that fine-tunes the skills instead of using a corrective policy, using the same reward as ASC.
    This ablation studies if fine-tuning could be used instead of a corrective policy to adapt pre-trained experts in out-of-distribution states.
    \item Mono-RL: A single monolithic neural network policy, commanding both the arm and base actions.
    Mono-RL is pre-trained using DAgger, same as ASC, and fine-tuned using RL with the same reward as ASC.
    \item \nodagger: An ablation of ASC which does not pre-train the coordination policy using DAgger. 
\end{itemize}

Tab.~\ref{table:success_table} shows a quantitative comparison over 1500 \task episodes in 79 different furniture layouts in ReplicaCAD (mean success and standard error).
Learned skill coordination and correction in ASC outperforms Seq-skills ($94.9\% \pm 0.6$ versus $87.8\% \pm 0.8$), by avoiding hand-off errors in pick/place.
\nocorrective\ is unable to generalize to states which are out-of-distribution for pre-trained skills, and hence performs slightly worse than ASC ($92.3\% \pm 0.7$ for No-corrective versus $94.9\% \pm 0.6$ with ASC). 
\finetuning\ suffers from catastrophic forgetting, and performance deteriorates significantly, even with a small learning rate ($43.9\% \pm 1.2$).
Mono-RL also performs significantly worse, despite being pre-trained using DAgger, due to the sparsity of the reward function ($35.8\% \pm 1.2$ using Mono-RL).
This is consistent with findings in~\cite{szot2021habitat} that monolithic RL needs well-tuned rewards to learn complex tasks.
\nodagger\ performs the worst of all baselines ($5.6\% \pm 1.0$), showing the importance of pre-training the coordination policy with DAgger when using sparse rewards.
Each component of ASC is important for robust performance. Skill coordination outperforms skill sequencing, and the corrective policy adds robustness by adapting to states that are out-of-distribution for the pre-trained skills without catastrophic forgetting.

\xhdr{Coordination vs. corrective policy.}
In Tab.~\ref{table:success_table}, we see that adding the coordination policy to Seq-skills improves performance significantly (row 3${\rightarrow}$2, \til$12.6\% {\uparrow}$ across columns), and adding the corrective policy leads to further improvement (row 2${\rightarrow}$1, \til$6.4\% {\uparrow}$ across columns).
Learning a coordination policy enables better hand-off between navigation and pick/place skills, over sequentially executing skills. The corrective skill further improves performance by adapting to out-of-distribution states, not seen during basic visuomotor skill training.
 For example, the corrective policy can move the robot away from the navigation target to pick/place an object.

\label{sec:bd_comparison}
\begin{table}[t]
    \centering
    \resizebox{0.95\columnwidth}{!}{
        \begin{tabular}{ccccc}
\toprule
& Easy Nav & Medium Nav & Difficult Nav & Picking \\
\midrule

ASC skills & 100\% & 100\% & 100\% & 100\% \\
BD API & 100\% & 0\% & 0\% & 33\% \\
\bottomrule
        \end{tabular}
    }
\caption{
The BD API can navigate to unobstructed goals (\textit{Easy Nav}), but fails to circumvent obstacles (\textit{Medium Nav}) or exit a room (\textit{Difficult Nav}). It grasped 33\% of objects placed in front of the robot, and cannot grasp occluded objects, or objects not in the camera view. ASC's learned skills significantly enhance the capabilities of Spot, succeeding in 100\% of cases. Success is reported over three navigation episodes per difficulty, and 18 pick episodes in the \textit{Lab}.
}
\label{table:bd_api_comparison}
\vspace{-0.2cm}
\end{table}
\subsection{Robustness to perturbations}
An assistive robot should be robust to perturbations such as a human walking in front of it, or slight movements of the object that it is picking.
Since ASC uses reactive visuomotor skills (and is not reliant on an outdated map), it is robust to such disturbances (see Fig. \ref{fig:perturbation}).
When faced with changes in environment layout or dynamic obstacles, ASC re-routes the robot to a new collision-free path to the goal.
Since ASC does not take the object position as input, it is also robust to movements of the target object.
If the object is moved while the robot is searching, ASC continues searching for the object.
On hardware, sometimes the robot becomes unstable when picking, prompting the BD API to move the robot away from the object to re-gain balance.
Such cases are out-of-distribution, as ASC is not trained on such disturbances, but ASC re-activates $\pi_{nav}$ and moves the robot back towards the object, before picking the target object. In these cases, the importance of the coordination policy is evident, as it considers the most recent observations to reactively decide whether to re-invoke navigation or pick/place skills.
On the other hand, Seq-skills and \seqskillsmap\ are not robust to such disturbances, as they follow a pre-defined sequence of skills, and do not take current observations into account when deciding which skill to activate, ultimately failing to pick the target object.
Robustness to such disturbances is achieved without any explicit training, emerging naturally from the design choices made in ASC.

\subsection{Comparing ASC skills and the Boston Dynamics API}

Boston Dynamics provides navigation, manipulation, and grasping APIs on Spot, which serve as the backbone for ASC.
This has two advantages: ASC is able to leverage controllers specifically designed and tuned for Spot by BD, and it avoids simulating costly and slow high-fidelity, low-level physics, but still enhances robot capabilities through learning in simulation.
Here, we show that ASC's learned skills significantly enhance the abilities of Spot, and are essential for consistently succeeding at long-horizon \task. 
ASC's learned navigation skill can navigate Spot to distant parts of the environment, while the BD navigation API can only navigate to unobstructed goals without a map.
As shown in Tab.~\ref{table:bd_api_comparison}, in the real world, both BD navigation and the learned navigation skill can reach unobstructed goals (\textit{Easy Nav}), but BD navigation fails when asked to circumvent an obstacle blocking its path (\textit{Medium Nav}), or reach goals in another room (\textit{Difficult Nav}).
For such challenging goals, the BD API requires a map of the environment, but can fail if layout of the environment changes~\cite{spotgraphnav}.
In contrast, the learned navigation skill can perform long-range navigation on Spot in all cases, and is robust to changes in environment layout.

The BD grasp API typically fails to grasp the target object if it is partially occluded, as it may opt to grasp the occluding object instead.
In contrast, ASC's learned pick skill searches for objects if they are not initially visible, and moves the arm such that the target object is at the center of the gripper camera's image, before calling the BD grasp API.
Tab.~\ref{table:bd_api_comparison} shows that ASC's learned pick skill is able to search for and pick objects in 18/18 episodes, including when occluded, while BD grasp only succeeds in 6/18 episodes, when objects are clearly visible.
However, despite ASC's enhancements, the BD grasp may sometimes fail to pick an object due to hardware instability or camera latency.
In such cases, ASC re-activates the pick skill, which searches for the object, and re-grasps.

%% file: 7_conclusion.tex
\section{Limitations and Conclusion}
\looseness=-1
In this work, we present Adaptive Skill Coordination (ASC), an approach that coordinates pre-trained skills and adapts them based on observations.
ASC is deployed zero-shot on the Spot robot for \task in eight unseen real-world environments, and achieves near-perfect performance in quantitative experiments in two environments, without the need for detailed maps with obstacle layouts or precise object locations.
In comparison, sequential skill-chaining suffers from hand-off errors and inability to recover from disturbances.
Despite being trained entirely in simulation, ASC is robust to real-world disturbances such as dynamic obstacles, changes to environment layout, and target objects being moved mid-episode.

We see several limitations and areas for future work. First, while ASC is a reasonably general approach, our experiments are limited to \task of small objects from open receptacles.
Future work involves studying generalization to a larger number of skills in order to complete more complex mobile manipulation.
Second, while our task and goal specification is consistent with community benchmarks~\cite{batra2020rearrangement, perille2020benchmarking}, it can be improved by allowing goals to be specified with natural language instead.
Finally, our experiments are conducted in static environments where the robot is the only active agent; future work involves studying assistive mobile manipulation in multi-agent and dynamic environments.

%% file: 8_appendix.tex
\section{Appendix}
The appendix is structured as follows:
\begin{itemize}
    \item Real-world and simulated environments and observations (\ref{sec:obs})
    \item Network architectures (\ref{sec:network_architectures})
    \item Training and hardware evaluation time(\ref{sec:train_time})
    \item Boston Dynamics APIs (\ref{sec:bd_api})
    \item Baselines and ablations (\ref{sec:baselines})
    \item DAgger pre-training (\ref{sec:dagger})
    \item Object detection pipeline (\ref{sec:obj-detect})
    \item Spot robot hardware (\ref{sec:spot-robot})
    \item Additional robustness experiments (\ref{sec:add-robust})
\end{itemize}

\subsection{Real-world and simulated environments and observations}
\label{sec:obs}
\begin{figure*}
    \centering
    \includegraphics[width=0.9\textwidth]{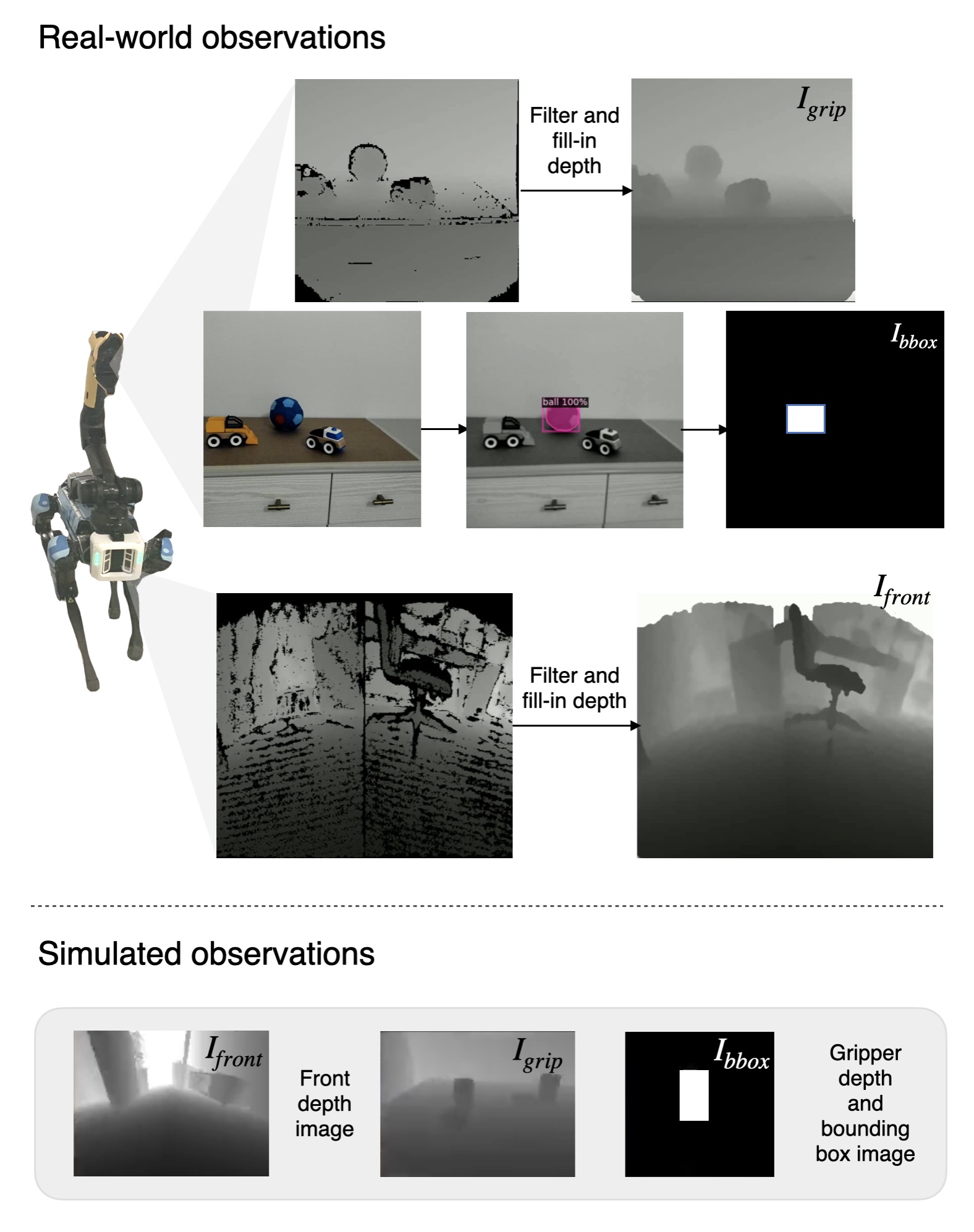}
    \caption{\small Simulated and real-world visual observations. The visual observations consist of:
(1) two front depth images from the robot, concatenated to make a single image $I_{front}$;
(2) a depth image from the gripper of the robot $I_{grip}$;
(3) a bounding box image around the target object, if detected by an object detector operating on the gripper camera $I_{bbox}$. To minimize the sim-to-real gap, we de-noise the real depth images using a fast depth completion algorithm that leverages classical image processing \cite{ku2018defense}. }
    \label{fig:real-world-obs}
\end{figure*}

Fig.~\ref{fig:real-world-obs} provides a visual comparison between the observations ASC uses within simulation and those that it uses when deployed on a real robot. In order to reduce the gap in appearance between the images from real-world depth cameras and those that the robot observed within simulation during training, we denoise the real depth images using the fast depth completion algorithm by Ku et al. \cite{ku2018defense}.

\begin{figure*}
    \centering
    \includegraphics[width=0.9\textwidth]{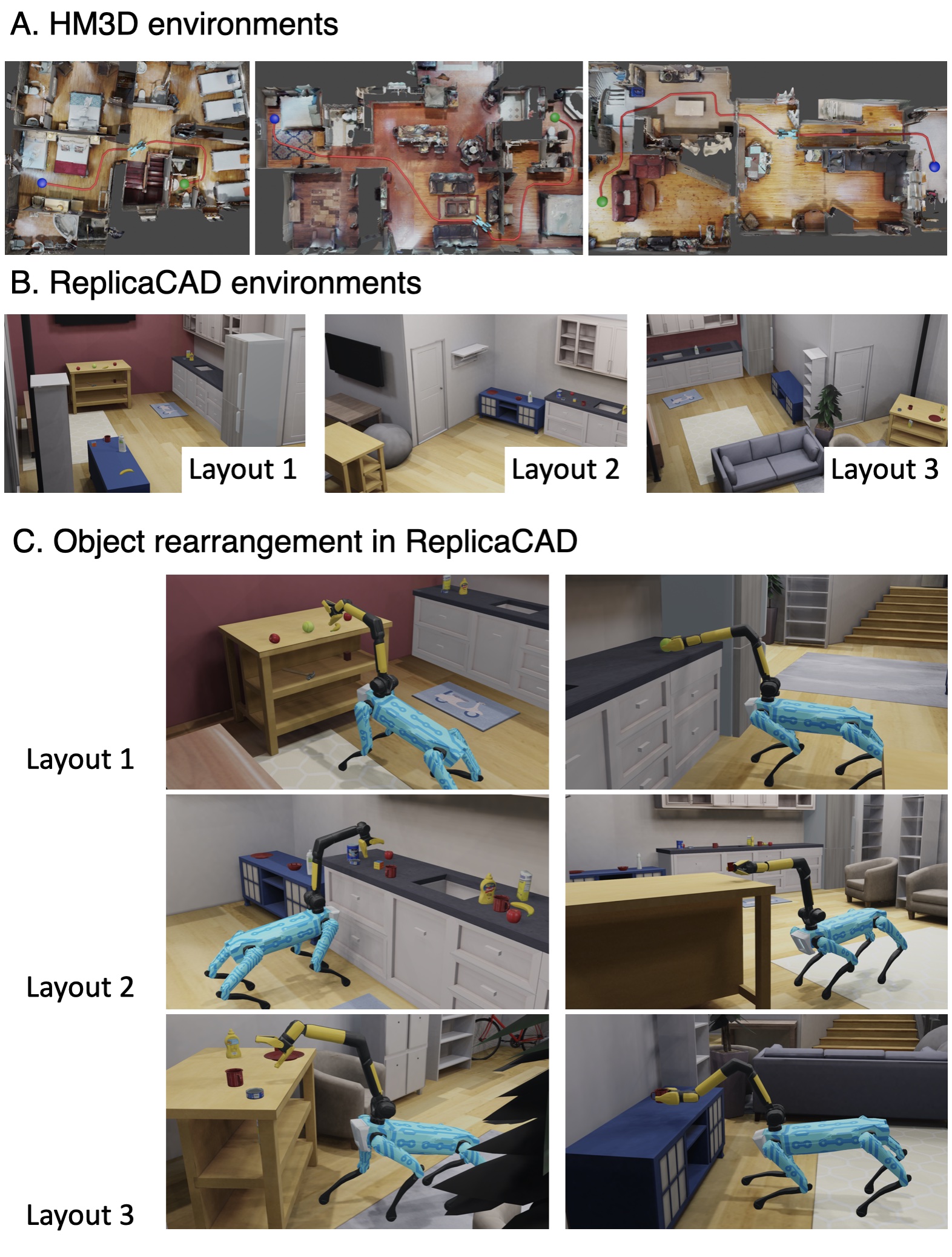}
    \caption{Simulation environments used during training. The navigation skill is trained on 1000 scans of the HM3D dataset (A), while the pick and place skills are trained in 104 layouts of the ReplicaCAD apartment (B). The coordination and corrective policies are trained on object rearrangement episodes in 104 different layouts of the simulated ReplicaCAD apartment (C).}
    \label{fig:sim-env}
\end{figure*}

Fig.~\ref{fig:sim-env} shows the different simulation environments used in the training of ASC. ASC can leverage and fully retain the experience gathered on different datasets simulated within Habitat.
The navigation skill is trained on the HM3D dataset \cite{ramakrishnan2021hm3d}, while the pick and place skills are trained using the ReplicaCAD \cite{szot2021habitat} and YCB datasets \cite{ycbdataset}.
The HM3D dataset contains 1,000 high-quality 3D scans of real-world indoor environments, making it an excellent source of data for training navigation skills (Fig.~\ref{fig:sim-env}A).
We train our navigation skill on 800 of these scans and use the rest as a validation set to evaluate the performance of the navigation expert and select the best checkpoint to use for our navigation skill. 
However, since the HM3D scans are static, without interactive objects or furniture, they are not suitable for learning pick and place skills.
Instead, we use the ReplicaCAD dataset, which has interactive receptacles such as cabinets, drawers, and refrigerators that open/close, for learning the pick, place, coordination, and corrective policies. 
The robot is tasked with picking, placing or rearranging one of 13 objects from the YCB dataset on one of 4 pieces of furniture (receptacles) in 104 different layouts of the simulated ReplicaCAD apartment.
Fig.~\ref{fig:sim-env}B shows 3 different layouts on ReplicaCAD, and Fig.~\ref{fig:sim-env}C shows Spot rearranging an object in each layout.

\subsection{Network architectures}
\label{sec:network_architectures}
The three basic visuomotor skills and coordination and corrective policies (henceforth collectively referred to as `policies') have similar network architectures. All policies, except place, start with a visual encoder which takes raw images $I$ as input, and returns an image embedding of size 512 $s_I$: $s_I = \phi(I)$. The coordination and corrective policies use two visual encoders (one from the trained navigation skill, another from the trained pick skill), whose outputs are concatenated. A visual encoder consists of three 2D convolutional layers, with output channel sizes of \{32, 64, 32\}, and kernel sizes of \{(8,8), (4, 4), (3, 3)\}, respectively. The output feature map of these convolutional layers is flattened and passed through a linear layer of output size 512 to obtain the image embedding $s_I$. Next, $s_I$ is concatenated with the rest of the non-visual observations (depending on the policy), and passed through a 2-layer multi-layer perceptron (MLP) with 512 hidden units (for both layers). The output of the MLP, $s_{MLP}$, is passed through a gated recurrent unit (GRU) \cite{cho2014properties} with 1 hidden layer of size 512, and an output layer of size 512. The output of the GRU constitutes the encoded state $s_{enc}$ of the robot, with the recurrent unit accounting for partial observability by storing temporal information. All hidden layers have ReLU activations. The input and output dimension of the GRU network depends on the policy, and is detailed in Table S1.

All policies have four outputs -- (1) the current encoded state $s_{enc}$, (2) the mean $\mu$ and (3) standard deviation $\sigma$ of a diagonal Gaussian distribution, and (4) an estimate of the value of the current state. The encoded state $s_{enc}$ outputted by the GRU layer is passed through three parallel linear output layers (heads); each of these provide the other three outputs (outputs 2-4). The output heads for $\mu$ and $\sigma$ are the size of the action (depends on the policy, listed in Table S1). For a policy $\pi$ given observations $o$, an action $a$ is sampled as: $a \sim \mathcal{N}(\mu, \sigma \textbf{I})$, where $\mu$ is the mean, $\sigma$ is the standard deviation, and $\textbf{I}$ is the identity matrix. The output layer that predicts the action mean $\mu$ is passed through a $tanh$ function, and the variance $\sigma$ is clipped between $10^{-6}$ and $1$.
The third head is a one-dimensional critic head that estimates the value of the current state $s_{enc}$, used during reinforcement learning to calculate the actor loss for DDPPO \cite{wijmans2020ddppo}. The output layer that estimates the value has a linear activation.  

\begin{table}[]
    \centering
    \begin{tabular}{c|c|c}
         Policy & Input dim & Output dim  \\
         \hline
         Navigation & 512 + 3 & 2 \\
         Pick & 512 + 4 & 4 \\
         Place  & 7 & 4 \\
         Coordination & 512 + 512 + 10 & 3 \\
         Corrective & 512 + 512 + 10 & 6 \\
    \end{tabular} \\
    \vspace{0.5cm}
    Table S1: Input and output (action) dimensions of each policy.
\end{table}

\subsection{Training and hardware evaluation time}
\label{sec:train_time}
For training the navigation skill, we utilize 8 GPUs (Nvidia Titan Xp, or similar) with 24 workers each, for a total of 192 workers collecting experience in parallel. We train for 190 million steps for about 14 hours. For the pick and place skills, we utilize 1 GPU with 32 workers and train for 18 million time steps, which takes approximately 22 hours. For training the coordination policy using DAgger, we also use 1 GPU with 32 workers, and train for 1.3 million steps for about 3 hours. Finally, for training both the coordination and corrective policies using deep reinforcement learning, we use 8 GPUs with 8 workers each, and train for 150 million steps for about 45 hours.

\subsection{Boston Dynamics APIs}
\label{sec:bd_api} 
Spot comes with mature navigation and manipulation APIs provided by Boston Dynamics for controlling the robot. These APIs form the backbone hardware controllers of ASC. Here we describe the APIs used by ASC in detail. For more details, we refer to the BD Spot SDK \cite{SpotSDK}. Our supplementary video (available at \href{\ascURL}{\ascURLtext}) compares the performance of these APIs against ASC's learned skills, and shows that ASC significantly enhances the capabilities of Spot through learning in simulation.

\xhdr{Navigation API.}
The navigation API we use to move the robot takes in linear and angular velocity commands for the base of the robot, and moves the the robot to achieve the commanded velocity. If an obstacle is blocking the robot's path, this API stays in-place, and does not move the robot around it. As a result, it can reach unobstructed goals, but has difficulty reaching obstructed and distant goals, for example in another room. However, ASC's learned navigation skill can perform long-range navigation using the velocity tracking API, enhancing the robot's capabilities.

For the Seq-skills-map baseline, we use the NavGraph API provided by BD, which can navigate to distant goals, but requires a pre-built map. This involves manually driving the robot to record various paths throughout the environment in order to build a map, before navigating. The map is built once, and assumed static. As a result, unlike our learned navigation skill, the NavGraph API is susceptible to changes in the environment; if the environment changes significantly between building a map and testing (for example, if a previously blocked path is opened, or a previously unblocked path is blocked), the map is not updated, causing inefficient paths or even failure. In contrast, our learned navigation skill requires no pre-built map, and can re-route the robot around dynamic obstacles, and is thus robust to such changes in the environment.

The navigation API also provides an egomotion estimate, or the robot's current relative distance and heading from where it started. Retrieving this data does not require any prior knowledge or exploration of the environment; the estimate is computed using onboard odometry (no external cameras or sensors) provided by Boston Dynamics.

\xhdr{Grasp API.}
The grasp API provided by BD can grasp unknown, arbitrary objects, if provided a pixel lying on the object in an image from one of the robot cameras. For ASC, we use the robot's gripper camera image, and send the center of the target bounding box to the grasp API. The grasp API then executes whole-body motion planning to grasp the object, and typically succeeds as long as the object is visible, unoccluded, and close to the gripper. These conditions are all met when ASC's pick skill is used to position the arm first. If the BD grasp API fails to grasp an object, it typically returns an error (for example, when manipulation planning fails). In such cases, ASC can re-activate the pick policy and resume searching for objects again.

\xhdr{Manipulation API.}
The manipulation API is given a target pose of arm joints, as well as a specified amount of time in which to reach this target pose. In our experiments, we give the API 0.5 seconds. At each time step, ASC moves each joint up to 10$^{\circ}$, for a maximum angular velocity of 20$^{\circ}$ per second for each joint. In the future, we could experiment with sending trajectories, or higher frequency velocity commands, both of which are supported by the manipulation API.

\begin{figure*}[t!]
\centering
    \includegraphics[width=1.0\textwidth]{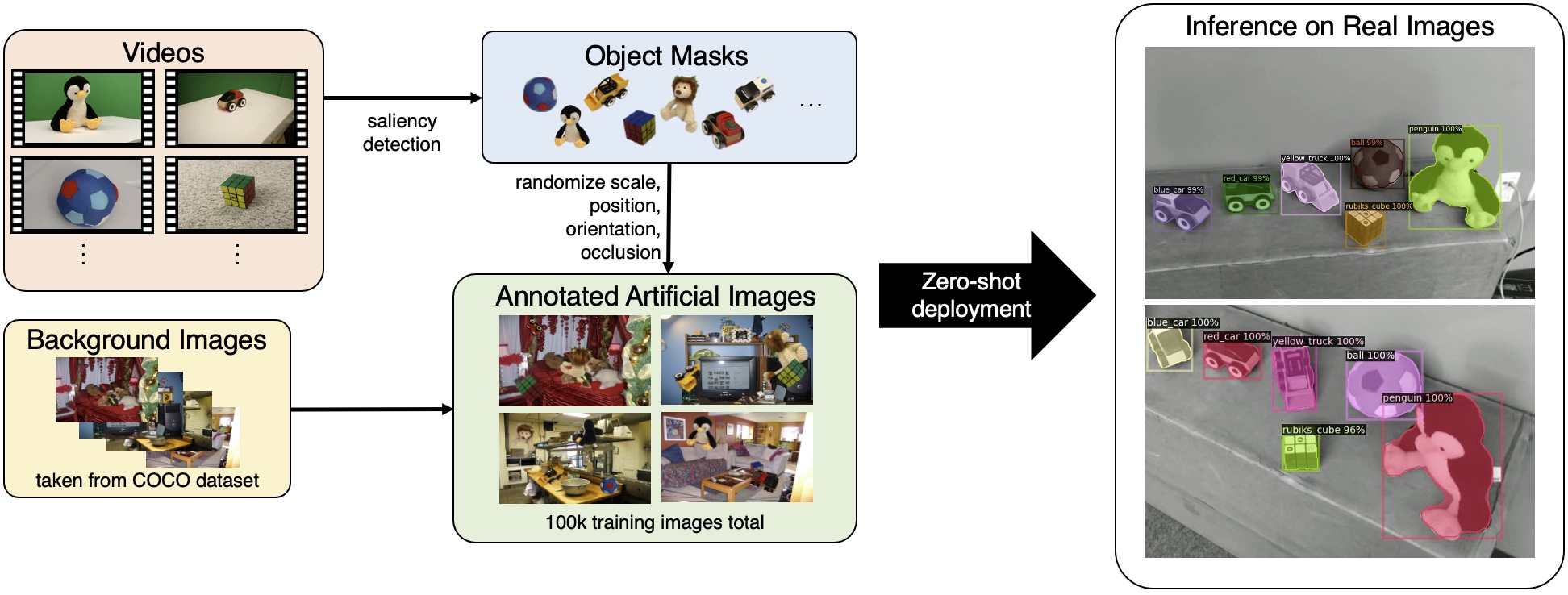}
    \caption{
        Training pipeline for the real-world object detector. We generate 100k automatically annotated artificial images for training an object detector using only unlabeled videos of objects and images from the COCO dataset. The detector is then deployed zero-shot to images taken from the real robot's gripper camera.
    }
    \label{fig:apdx_obj_det_training}
\end{figure*}

\subsection{Baselines and ablations}
\label{sec:baselines}

For the metrics depicted in Tab.~\ref{table:success_table}, we train three policies from three random seeds, and compute mean and standard error for each approach. Policy weights used in evaluation are selected based on their performance on a held-out validation set of 1500 mobile manipulation episodes. All approaches utilize the same superset of skill observations $o$, and the same action space $(v, \omega, \Delta q_{arm})$. 

\xhdr{More details on Seq-skills.}
\textit{Seq-skills} behaves in the following manner: if the robot is not within the success radius and heading (0.3m and 5$^\circ$) of the navigation goal $(x, y, \theta)$, its actions are determined by the navigation skill; otherwise, they are determined by the pick skill if the robot hasn't picked the object yet, or the place skill if it has. The navigation target is reset every time the agent successfully picks or places the object. It is changed either to the location of the target place receptacle once the target object has been picked, or the next clutter receptacle if the object has been placed.

\xhdr{More details on Mono-RL.}
\textit{Mono-RL} utilizes a policy architecture that consists of two visual encoders (one for $I_{front}$, another for $I_{grip} + I_{bbox}$) and a downstream MLP-GRU policy with the same architecture described in Appendix \ref{sec:network_architectures}. Its action distribution outputs both base and arm actions.
Like ASC, we utilize DAgger pre-training (see Appendix \ref{sec:dagger} for how labels are generated for Mono-RL) to warm-start the weights of \textit{Mono-RL} before using deep RL with the sparse long-horizon reward function (see Section \ref{subsec:skill_coor_and_corr}). 
However, during RL, due to the sparsity of the reward and the lack of (frozen) pre-trained skills, \textit{Mono-RL} is unable to learn the task, despite the DAgger initialization.

\xhdr{GRU vs. MLP.}
We also experiment with removing the recurrent components of the coordination and corrective policies.
We see a decrease in performance when the GRU layers are replaced by MLP layers (from 94.9$\pm$0.6 to 93.3$\pm$0.7, across three seeds).
We also evaluate both GRU-based and MLP-based versions of ASC on a harder dataset, in which the navigation goals for both picking and placing locations are set farther away from the corresponding target.
In the normal dataset, the picking or placing target can be up to 0.45 m and 30$^{\circ}$ away from their corresponding navigation goal, whereas they can be up to 1 m and 60$^{\circ}$ away in the harder dataset.
On the harder dataset, we see a larger decrease in performance with using an MLP instead of a GRU (from 65.4$\pm$1.2 to 60.7$\pm$1.3). These results are summarized in the table below:

\begin{table}[ht]
    \centering
    \begin{tabular}{c|c|c}
         Dataset & GRU-based & MLP-based  \\
         \hline
         Normal & 94.87\tiny{$\pm$}0.57  & 93.27\tiny{$\pm$}0.65 \\
         Hard & 65.4\tiny{$\pm$}1.23 & 60.73\tiny{$\pm$}1.26 \\
    \end{tabular} \\
    \vspace{0.5cm}
\end{table}

\pagebreak

Lastly, we run an ablation over the reward weights used in the sparse long-horizon reward function in Section \ref{subsec:skill_coor_and_corr} and analyse the sensitivity of ASC's performance to these changes. We use 3 sets of weights, and find the performance of ASC remains similar across them.

\begin{table}[ht]
    \centering
    \begin{tabular}{c|c}
         Weights & Performance  \\
         \hline
         $[10, 0.03, 0.03, 0.0003]$ & 94.87\tiny{$\pm$}0.57  \\
         $[5.0, 0.01, 0.01, 0.0001]$ & 92.8\tiny{$\pm$}0.67  \\
         $[15, 0.05, 0.05, 0.0005]$ & 93.5\tiny{$\pm$}0.64
    \end{tabular} \\
    \vspace{0.5cm}
\end{table}

\subsection{DAgger pre-training}

\label{sec:dagger}
In DAgger training, a `student' policy is trained using supervised learning to behave like a `teacher' policy that provides the target actions at each time step. However, unlike behavior cloning, the actions applied in the environment during training are the student actions, reducing the distribution shift when the student policy is deployed without the teacher.
For ASC, we use DAgger to train the gating network $\pi_z$ using the same logic that drives the \textit{Seq-skills} (see Appendix \ref{sec:baselines}) baseline. $\pi_z$ is provided with one-hot encoded vectors that identify which skill it should use at each step. Specifically, if the robot is not within the success condition for navigation, the action label that $\pi_z$ is trained to output would correspond to activating just the navigation skill; otherwise, if the robot hasn't picked the object yet, the label corresponds to activating just the pick skill, else the label corresponds to activating just the place skill. There are no labels for the corrective policy, hence it is not warm-started using DAgger. Both coordination and corrective policies are then fine-tuned using deep reinforcement learning.

For pre-training the \textit{Mono-RL} baseline using DAgger (before deep reinforcement learning fine-tuning), the labels are the teacher base or arm actions, instead of one-hot encoded vectors. Once the correct skill for the current time step is identified by the \textit{Seq-skills} logic, we use the output of that skill as teacher actions. Because \textit{Seq-skills} only controls either the base or the arm (but not both) at each time step, the part of the action label corresponding to the unused portion of the action space is set to zero.

We find that pre-training with DAgger substantially improves success rate compared to using RL alone for both ASC and \textit{Mono-RL}. 

\subsection{Object detection pipeline}
\label{sec:obj-detect}
For our quantitative experiments, we use the Mask R-CNN~\cite{he2017mask} object detector. To train it, we generate a dataset of automatically annotated images by overlaying object contours on to background images from the COCO dataset \cite{lin2014microsoft} (see Fig.~\ref{fig:apdx_obj_det_training}). To extract the object contours, we rely on a U\textsuperscript{2}-Net \cite{u2net} model that leverages ReSidual U-blocks (RSU) in its architecture. The U\textsuperscript{2}-Net is applied to a video of each object, in which the object was seen from a wide variety of viewing angles, which gives us segmented images of the object from different angles. We do not train or fine-tune this model, and instead use a pre-trained set of weights released by the authors. Next, inspired by \cite{frasier_robocup}, we randomly resize, rotate, flip and superimpose these extracted object contours on 100,000 images from the COCO object detection dataset, to create an automatically labeled dataset. The dataset contains automatically labeled segmented masks of objects, as well as distractor objects originally present in the COCO dataset, and can then be used for training an object segmentation model such as Mask R-CNN \cite{he2017mask}. 
We use the Mask R-CNN implementation from the Detectron2 library of detection algorithms by Meta AI \cite{wu2019detectron2}.
We use a backbone that leverages ResNet-101 \cite{he2016deep} and feature pyramid networks (FPNs) \cite{lin2017feature}.
At the start of training, the network is initialized with pre-trained weights that are available in Detectron2's model zoo.
We find that converting all training images to grayscale and 
Upon deployment for real world experiments, our Mask R-CNN model takes about 200 milliseconds to generate object detections using the gripper's RGB images.

\subsection{Spot robot hardware}
\label{sec:spot-robot}
ASC is deployed zero-shot to the real-world on a Spot robot \cite{Spot}.
Spot is equipped with five D430 stereo cameras on its base (ASC only uses the two front-facing cameras).
In addition, it has a 6 degrees-of-freedom arm and a jaw gripper equipped with an RGB-D camera.
The robot comes with mature low-level controller APIs (details in Appendix \ref{sec:bd_api});
ASC makes use of Spot's egomotion estimates, its velocity controllers for executing navigation actions, and its grasp API to pick the target object.
All ASC policy outputs are sent at 2 Hz to the robot, using a computer equipped with an RTX 3070 GPU, while observations from the robot are updated at about 12Hz asynchronously.
This allows the policies to receive the most recent observation as input to reason about robot actions, and hence be reactive to real-world disturbances like moving obstacles, objects, etc.

\subsection{Additional robustness experiments}
\label{sec:add-robust}
\begin{figure}[h!]
    \centering
    \includegraphics[width=0.5\textwidth]{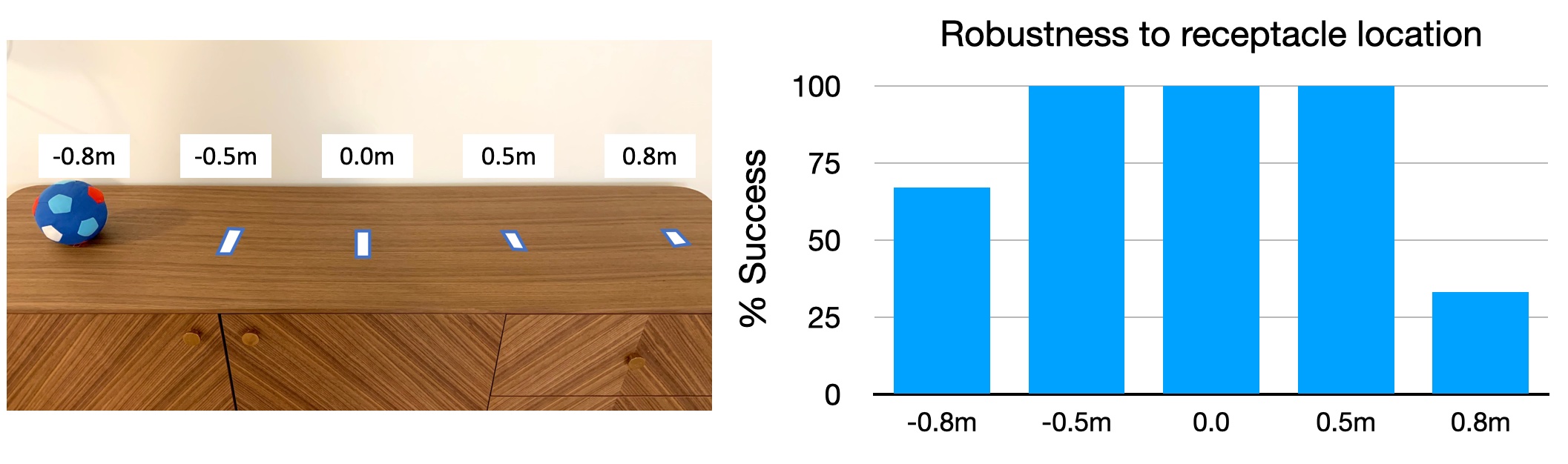}
    \caption{Robustness to object location}
    \label{fig:robust-obj}
\end{figure}
A natural form of disturbance in object rearrangement is the location of the object on a receptacle. For example, the object could be located near the center of the receptacle (e.g.,~`hall table'), making it easy to find and grasp it, or it might be at the edge of the receptacle, and the robot has to search for it before grasping. Since the rearrangement task provides only the receptacle location and not the precise location of the object, any approach must be robust to such environmental disturbances. Fig.~\ref{fig:robust-obj} shows an experiment where we move the object further and further away from the receptacle center and measure ASC's performance at successfully grasping and placing the object. We observe that ASC can successfully grasp the object in 9/9 episodes for small perturbations when the object is placed $< \pm 0.3$m away from the receptacle center, and 3/6 times for large perturbations when the object is $\pm 0.6$m away from the center (Tab.~\ref{table:success_table}). Robustness to small perturbations is achieved without any explicit training for perturbations, and emerges from the design choices made in ASC. However, ASC's performance on larger systematic disturbances can be further improved by training ASC with disturbances in simulation, as is common practice in robot learning works like \cite{tan2018sim, miki2022learning, peng2018sim}.